%% file: main.tex
\documentclass{article}

\usepackage{soul_style}
\usepackage[export]{adjustbox}
\usepackage[utf8]{inputenc} %
\usepackage[T1]{fontenc}    %
\usepackage{newunicodechar}
\usepackage{hyperref}       %
\usepackage{xcolor}
\usepackage[normalem]{ulem} %
\hypersetup{
    colorlinks=true,      %
    linkcolor=blue,      %
    urlcolor=blue,       %
    citecolor=blue,      %
    linkbordercolor=blue, %
    urlbordercolor=blue,
    citebordercolor=blue,
    pdfborderstyle={/S/U/W 1}, %
}

\usepackage{float}
\usepackage{url}            %
\usepackage{booktabs}       %
\usepackage{amsfonts}       %
\usepackage{nicefrac}       %
\usepackage{microtype}      %
\usepackage{lipsum}		%
\usepackage{graphicx}
\usepackage{natbib}
\usepackage{doi}
\usepackage{amsmath}
\usepackage{amssymb} %
\usepackage{xspace}
\usepackage{enumitem}
\usepackage{multirow}
\usepackage{subcaption} 
\usepackage{makecell}
\usepackage{hyperref, cleveref}
\usepackage{pifont}
\usepackage[inkscapelatex=false]{svg}
\usepackage{caption}
\usepackage{wrapfig}
\usepackage{algorithm}
\usepackage{algorithmic}
\usepackage{threeparttable}
\usepackage{overpic}

\setlist[itemize]{leftmargin=*}
\setlist[enumerate]{leftmargin=*}
\setlist[description]{leftmargin=*}

\newcommand{\soulxflashtalk}{SoulX-FlashTalk\xspace}

\definecolor{midnightgreen}{rgb}{0.0, 0.29, 0.33}

\title{\textnormal{\soulxflashtalk: Real-Time Infinite Streaming of Audio-Driven Avatars via Self-Correcting Bidirectional Distillation}}

\author{
\normalfont Le Shen\textsuperscript{1,2,\ddag,}\thanks{Equal contribution. $^\ddag$Work done during an internship at Soul AI Lab. le.shen@mail.dhu.edu.cn, \{qiaoqian,yutan\}@soulapp.cn}, Qian Qiao\textsuperscript{1,}$^*$, Tan Yu\textsuperscript{1,}$^*$, \normalfont Ke Zhou\textsuperscript{1}, \\
Tianhang Yu\textsuperscript{1}, Yu Zhan\textsuperscript{1}, Zhenjie Wang\textsuperscript{2}, Dingcheng Zhen\textsuperscript{1}, Ming Tao\textsuperscript{1}, Shunshun Yin\textsuperscript{1}, Siyuan Liu\thanks{Corresponding authors. liusiyuan@soulapp.cn}   \\
    \textsuperscript{1}AIGC Team, Soul AI Lab, China \quad \textsuperscript{2}Donghua University\\
      \textbf{Project Page:} \href{https://soul-ailab.github.io/soulx-flashtalk/}{https://soul-ailab.github.io/soulx-flashtalk/}\\
}
\renewcommand{\headeright}{\raisebox{-0.2\height}{\includegraphics[height=1.8em]{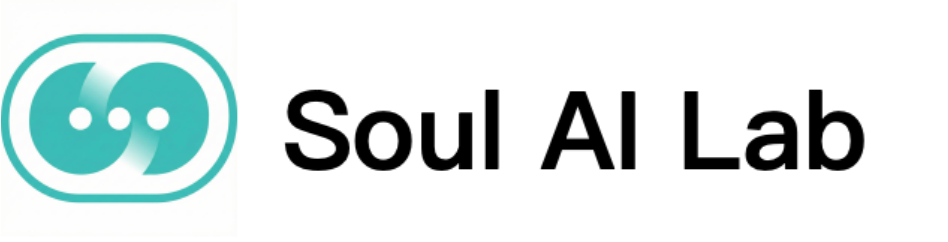}}}
\renewcommand{\shorttitle}{SoulX-FlashTalk Technical Report}

\usepackage{amsmath}
\usepackage{CJKutf8}
\begin{document}
\maketitle

\input{content_arxiv/0.abstract}
\input{content_arxiv/1.Introduction}
\input{content_arxiv/2.Method}
\input{content_arxiv/3.Experiments}
\input{content_arxiv/4.Conclusion}
\input{content_arxiv/6.Statement}
\input{content_arxiv/5.Contribution}

\bibliographystyle{unsrtnat}
\bibliography{references}  

\clearpage
\appendix

\end{document}

%% file: content_arxiv/0.abstract.tex
\begin{abstract}
Deploying massive diffusion models for real-time, infinite-duration, audio-driven avatar generation presents a significant engineering challenge, primarily due to the conflict between computational load and strict latency constraints. Existing approaches often compromise visual fidelity by enforcing strictly unidirectional attention mechanisms or reducing model capacity. To address this problem, we introduce \textbf{SoulX-FlashTalk}, a 14B-parameter framework optimized for high-fidelity real-time streaming. Diverging from conventional unidirectional paradigms, we use a \textbf{Self-correcting Bidirectional Distillation} strategy that retains bidirectional attention within video chunks. This design preserves critical spatiotemporal correlations, significantly enhancing motion coherence and visual detail. To ensure stability during infinite generation, we incorporate a \textbf{Multi-step Retrospective Self-Correction Mechanism}, enabling the model to autonomously recover from accumulated errors and preventing collapse. Furthermore, we engineered a full-stack inference acceleration suite incorporating hybrid sequence parallelism, Parallel VAE, and kernel-level optimizations. 
Extensive evaluations confirm that SoulX-FlashTalk is the first 14B-scale system to achieve a \textbf{sub-second start-up latency (0.87s)} while reaching a real-time throughput of \textbf{32 FPS}, setting a new standard for high-fidelity interactive digital human synthesis.
\end{abstract}

%% file: content_arxiv/1.Introduction.tex
\section{Introduction}
\input{figs/figure_teaser}
Diffusion Transformers (DiTs)~\citep{peebles2023scalable} serve as a scalable backbone for high-fidelity generative modeling. Building on this capability, recent large-scale models~\citep{gao2025wan,zhong2025anytalker,guo2024liveportrait,yang2025infinitetalk} have significantly advanced avatar generation, delivering cinematic quality, detailed micro-expressions, and natural full-body dynamics. However, deploying these models for real-time, infinite streaming remains a major engineering challenge. The primary conflict lies between the high computational cost required for high-fidelity generation and the strict low-latency demands of live streaming.

To achieve real-time performance, state-of-the-art approacheslike LiveAvatar~\citep{huang2025live} typically employ a \emph{bidirectional-teacher to unidirectional-student} distillation strategy, combined with multi-GPU parallel inference. 
While effective, this design requires a complex two-stage training pipeline comprising diffusion forcing and DMD~\citep{yin2024one}distillation. This process is computationally expensive and can require up to 27,500 training steps.
More critically, converting the student model to a strictly unidirectional architecture disrupts the spatiotemporal correlations inherent in video generation. This structural mismatch often results in rigid full-body dynamics and a loss of textural detail. 
Although methods like Self-Forcing~\citep{huang2025self} and Self-Forcing++~\citep{cui2025self} simulate autoregressive inference to mitigate error accumulation, applying these mechanisms effectively to a large-scale model under real-time constraints lacks a mature reference solution.

To address these challenges, we introduce \textbf{SoulX-FlashTalk}, a low-latency, real-time, audio-driven avatar framework built upon a 14B-parameter DiT. Compared with prior approaches, our method offers substantial advantages across model architecture design, training and distillation strategy, and inference-time optimization, enabling high-fidelity avatar generation while meeting the stringent latency requirements of live streaming.

First, regarding model architecture and generation quality, we diverge from the strictly unidirectional paradigm by adopting a \emph{bidirectional-teacher to bidirectional-student} strategy. We argue that in chunk-based streaming inference, future information \textit{within} the current chunk is available. Therefore, retaining the bidirectional attention mechanism for intra-chunk processing is not only feasible but beneficial. This design allows the student model to leverage local context for motion planning, significantly enhancing full-body coherence and detail. 
Crucially, maintaining high architectural alignment between the teacher and student simplifies the distillation task, avoiding the performance degradation caused by structural forcing.

Second, in terms of training efficiency, SoulX-FlashTalk streamlines the time-consuming training paradigm of LiveAvatar~\citep{huang2025live}.
This improvement is enabled by the architectural consistency described above, which allows us to adopt a lightweight yet effective training strategy.
\textbf{Latency-Aware Spatiotemporal Adaptation} adapts the model to operate effectively under lower spatial resolutions and shorter temporal horizons for real-time constraints. 
\textbf{Self-Correcting Bidirectional Distillation} further reduces inference steps and eliminates classifier-free guidance, while introducing a multi-step retrospective self-correction mechanism inspired by Self-Forcing++ to enable the model to recover from self-induced deviations and prevents collapse during long-horizon generation.

Finally, to meet strict real-time streaming requirements with a large-scale DiTs, we construct a full-stack inference acceleration solution. By utilizing xDiT's hybrid sequence parallelism (integrating Ulysses~\citep{jacobs2023deepspeed} and Ring Attention~\citep{liu2023ring}), we achieve an approximate $5\times$ speedup for single-step inference on an $8$ GPUs setup. We also address the 3D VAE bottleneck by introducing the slicing parallel strategy from LightX2V~\citep{lightx2v}, accelerating encoding/decoding by nearly $5\times$. Furthermore, we tailor our kernel implementations for the Hopper architecture to fully exploit its hardware capabilities by adopting FlashAttention3~\citep{shah2024flashattention}.
Combined with \texttt{torch.compile} optimizations, these measures maximize hardware utilization. 

As shown in Figure~\ref{fig:teaser}(a), our model converges to superior performance with only 200 distillation steps, yielding an efficiency improvement of approximately $23\times$ over LiveAvatar, which requires 27,500 steps. Meanwhile, the proposed system achieves a start-up latency of 0.87 s, about $3\times$ faster than prior baselines at 2.89 s. Together with these gains, this report presents comprehensive ablation studies of the proposed optimizations, providing practical guidance for the stable training and real-time deployment of large-scale DiTs.

%% file: figs/figure_teaser.tex
\begin{figure}[htbp]
  \centering
  \vspace{-20pt}
  \includegraphics[width=1.0\textwidth]{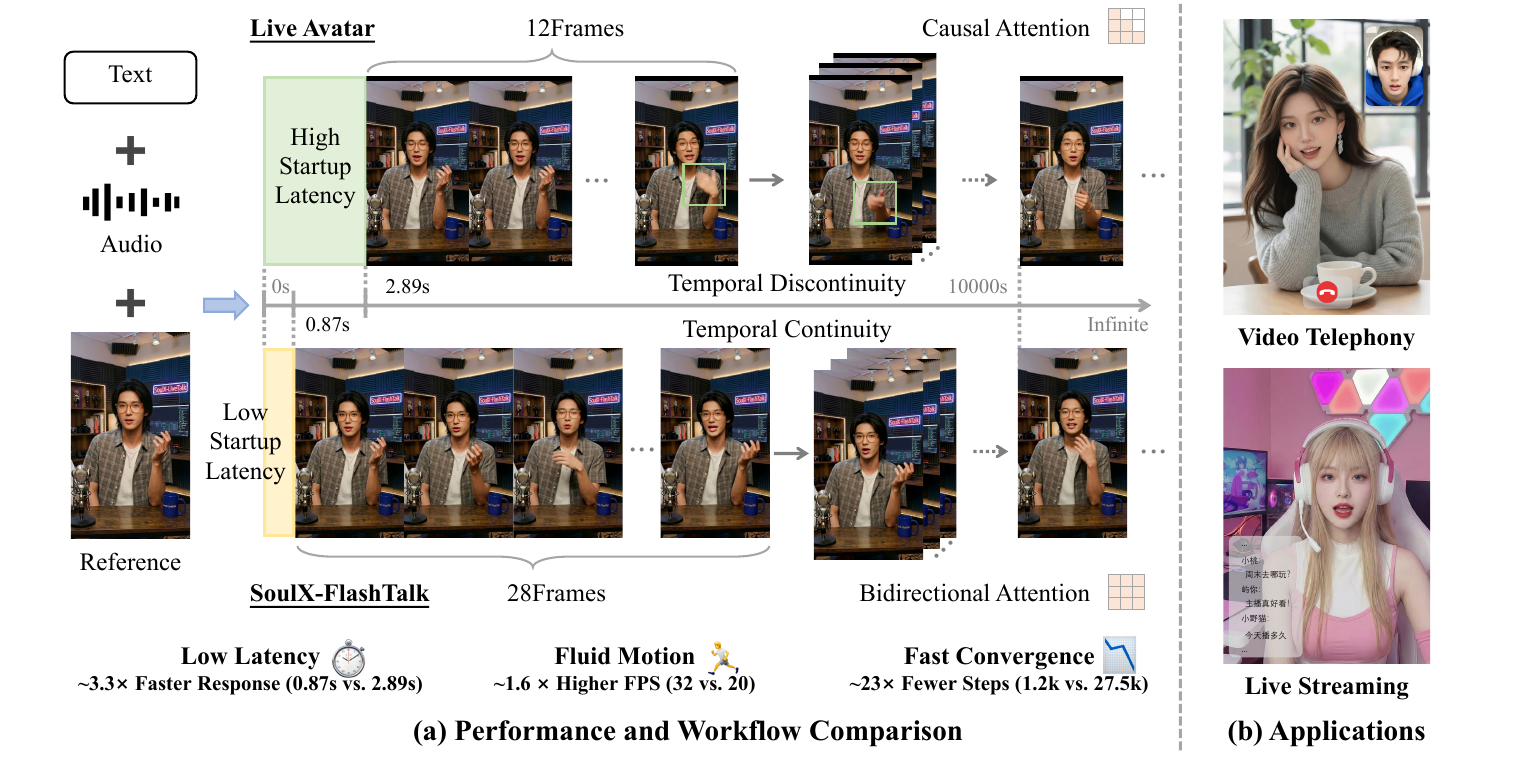}
  \vspace{-16pt}
  \caption{ Performance Overview and Applications of SoulX-FlashTalk.
(a) Performance and Workflow Comparison: Powered by our bidirectional streaming distillation, SoulX-FlashTalk converges in just 1.2k steps (reducing training costs by $\sim$\textbf{23$\times$} compared to LiveAvatar), achieves a 0.87s start-up latency ($\sim$\textbf{3.3$\times$} faster), and attains fluid motion at 32FPS ($\sim$\textbf{1.6$\times$} higher).
(b) Applications: Supports real-time interactions, including Video Telephony and Live Streaming.}
\label{fig:teaser}
\end{figure}

%% file: content_arxiv/2.Method.tex
\section{SoulX-FlashTalk}
\begin{figure*}[h]
    \centering
    \includegraphics[width=1.0\linewidth]{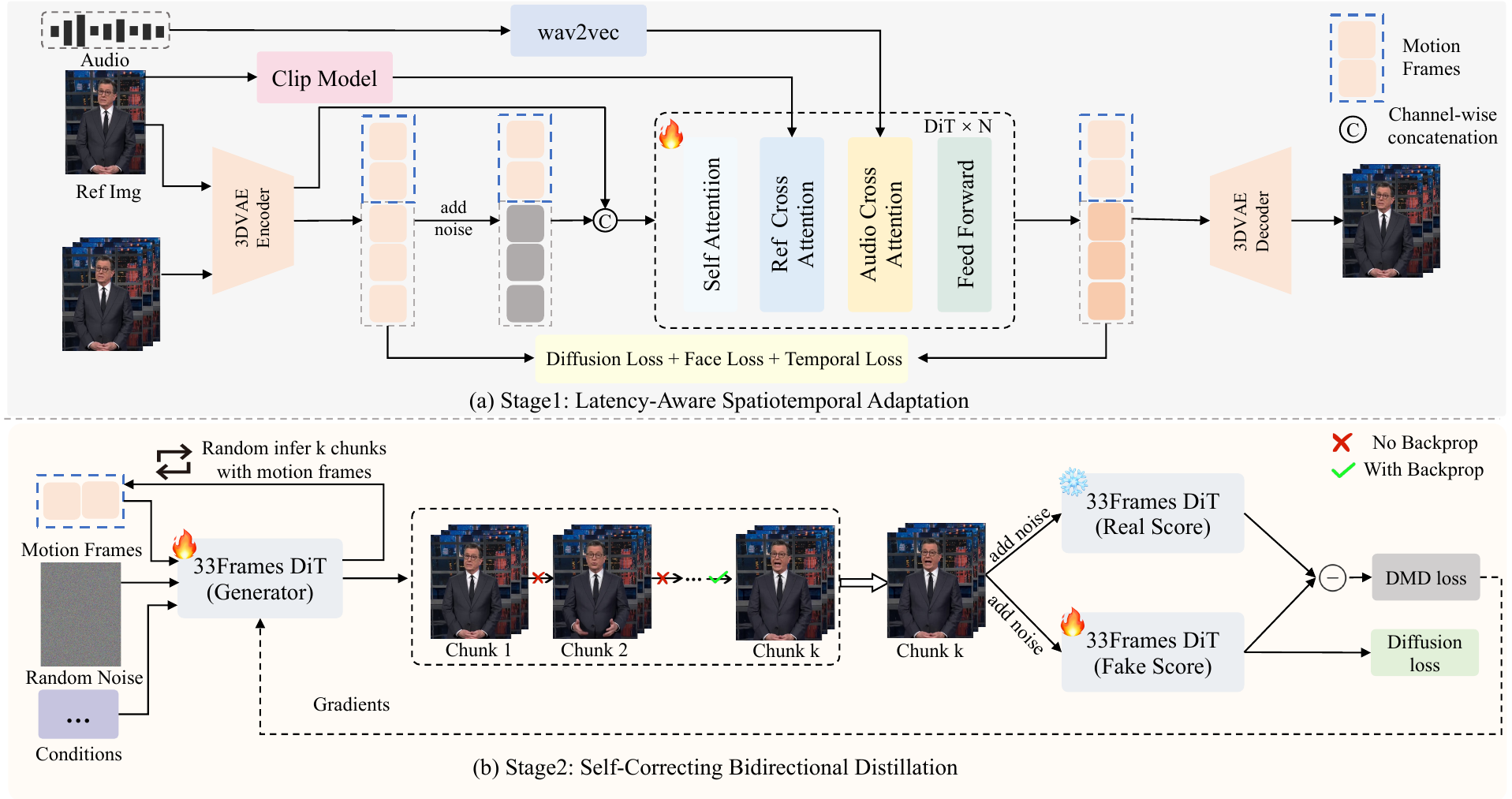}
    \caption{
    \textbf{Framework Overview.} 
    (a) Stage 1: Latency-Aware Spatiotemporal Adaptation, which adapts the model to operate effectively under lower spatial resolutions and shorter temporal frames to meet real-time constraints.
    (b) Stage 2: Self-Correcting Bidirectional Distillation, where the generator autoregressively synthesizes $k$ chunks conditioned on past motion frames, while the Real and Fake Score networks align data distributions through distillation losses.
    }
    \label{fig:method_overview}
\end{figure*}
\vspace{-5pt}
This section details the core methodology of SoulX-FlashTalk. As illustrated in Figure \ref{fig:method_overview}, our framework is built upon a 14B-parameter DiT and integrates a two-stage training pipeline with a full-stack inference acceleration engine. The training protocol progresses from Latency-Aware Spatiotemporal Adaptation phase to a Self-Correcting Bidirectional Distillation phase, designed to satisfy both high-fidelity generation and low-latency streaming requirements.

\subsection{Model Architecture}
The architecture derives from WAN2.1-I2V-14B~\citep{wan2025wan} and InfiniteTalk~\citep{yang2025infinitetalk}, comprising four primary components:

\textbf{3D VAE.} We utilize the WAN2.1 VAE for latent space compression to facilitate efficient high-resolution video generation. This module encodes video frames into compact latent representations, achieving a spatio-temporal downsampling factor of $4 \times 8 \times 8$ across temporal, height, and width dimensions.

\textbf{DiT-Based Generator.} The core generator adopts the DiT architecture~\citep{peebles2023scalable}. As shown in Figure.~\ref{fig:method_overview}(a), each DiT block incorporates a 3D Attention mechanism to model spatio-temporal dependencies. A unified cross-attention layer conditions the generation on reference images and textual inputs to preserve visual fidelity and provide semantic guidance. Furthermore, we integrate a dedicated audio cross-attention layer to inject speech-driven signals directly into the generation process.

\textbf{Conditioning Encoders.}
The model conditions generation on audio, text, and reference images. We employ a Wav2Vec model~\citep{baevski2020wav2vec} customized for Chinese speech to transform continuous audio signals into sequential embeddings. To ensure identity consistency, we extract semantic representations and visual features $f_{ref}$ from the reference image using CLIP~\citep{DBLP:conf/icml/RadfordKHRGASAM21} and the VAE encoder. For textual conditioning, we adopt umT5~\citep{DBLP:journals/jmlr/RaffelSRLNMZLL20} to support bilingual captions. These identity and textual conditions are injected via the cross-attention layers.

\textbf{Latent Input Formulation.}
For a given source video, we sample a clip $\mathbf{X}_{\text{clean}}$ of length $L_c$. The initial $L_m$ frames serve as motion frames $\mathbf{X}_{\text{mf}}$ to capture historical context, while the subsequent frames function as generation targets. A reference frame $\mathbf{X}_{\text{ref}}$ is randomly sampled from outside the clip boundaries. All inputs are encoded by the 3D VAE and assembled to form the DiT input $\mathbf{z}_{\text{in}}$:
\begin{equation}
\begin{aligned}
\mathbf{z}_{\text{in}} &= \mathbf{z}_{\text{noise}} \| \mathbf{z}_{\text{mask}} \| \mathbf{z}_{\text{cond}}, \quad \text{where } \mathbf{z}_{\text{mask}} = \mathbf{1}_{1} \oplus \mathbf{0}_{L_c-1}, \\
\mathbf{z}_{\text{noise}} &= \mathcal{E}(\mathbf{X}_{\text{mf}}) \oplus\psi(\mathcal{E}(\mathbf{X}_{\text{clean}}),t), \quad
\mathbf{z}_{\text{cond}} = \mathcal{E}(\mathbf{X}_{\text{ref}} \oplus \mathbf{0}_{L_c-1}),
\end{aligned}
\end{equation}
where $\oplus$ denotes frame-wise concatenation and $\|$ denotes channel-wise concatenation. $\mathcal{E}(\cdot)$ represents the 3D VAE encoder. The stream $\mathbf{z}_{\text{noise}}$ combines historical latents with noisy latents derived by applying the forward diffusion process $\psi(\cdot)$ to $\mathcal{E}(\mathbf{X}_{\text{clean}})$ at timestep $t$. The stream $\mathbf{z}_{\text{cond}}$ injects reference guidance, while $\mathbf{z}_{\text{mask}}$ identifies reference frames using a binary indicator. This composite input structure facilitates bidirectional interaction between historical motion context and the current generation target, allowing the model to correct accumulated errors using reference information.

\subsection{Model Training}
To satisfy real-time inference under strict latency constraints, we employ a two-stage training strategy. Latency-Aware Spatiotemporal Adaptation adapts the model to reduced spatial resolutions and shorter frame sequences, while Self-Correcting Bidirectional Distillation further reduces sampling steps and removes classifier-free guidance~\citep{ho2022classifier}. This two-stage procedure enables rapid model responses while preserving high generation quality.

\textbf{Stage 1: Latency-Aware Spatiotemporal Adaptation}

The high computational cost of the 14B-parameter DiT backbone poses a significant challenge for real-time applications. While the original InfiniteTalk model delivers high-quality results, its inference latency on standard hardware is too high for interactive streaming. Consequently, we adapt the model to function at reduced spatial resolutions and with shorter frame sequences.

Deploying the pre-trained model directly under these constrained settings results in poor feature alignment and reduced generation quality. We address this by performing a dedicated fine-tuning phase optimized for the target resolution and frame count. During this phase, we apply a dynamic aspect-ratio bucketing strategy to organize training samples efficiently, reducing data loss from padding or cropping. This process enables the 14B model to recover fine details and maintain identity consistency, even at lower resolutions.

\textbf{Stage 2: Self-Correcting Bidirectional Distillation}

Multi-step sampling and classifier-free guidance create significant computational overhead. We adopt the DMD framework~\citep{yin2024one} to compress the sampling steps and eliminate the need for guidance, enabling real-time streaming.

The framework aims to minimize the distributional discrepancy between the original teacher model and the distilled student model at each time step $t$, using the Kullback–Leibler (KL) divergence as the optimization criterion. The resulting training objective is formulated as:
\begin{equation}
\label{eq:3}
\nabla_\theta \mathcal{L}_{\text{DMD}} = -\mathbb{E}_{t, \mathbf{z}} \left[ \left( s_{\text{real}}(\psi(G_\theta(z),t), t) - s_{\text{fake}}(\psi(G_\theta(z),t), t) \right) \frac{\partial G_\theta(\mathbf{z})}{\partial \theta} \right],
\end{equation}
Where, $s_{real}(\cdot)$ is frozen to model the teacher distribution, while $s_{fake}(\cdot)$ is trainable and tracks the evolving student distribution. The student generator $G_\theta(\cdot)$ produces samples via few-step inference without classifier-free guidance. All components are initialized from the Stage-1 SFT model.

Standard DMD does not address error accumulation or identity drift in long-form videos. Inspired by Self-Forcing++~\citep{cui2025self}, we introduce Self-Correcting Bidirectional Distillation, which incorporates a multi-step retrospective self-correction mechanism to explicitly simulate error propagation during long-horizon generation. Specifically, the generator is required to autoregressively synthesize $K$
consecutive chunks, where each chunk is conditioned on the previously generated motion frame rather than the ground truth.

To balance computational efficiency and training stability, we further propose a \textbf{Stochastic Truncation Strategy}. Instead of synthesizing all $K$ chunks, we randomly sample a smaller value $k < K$ and generate only the first $k$ chunks. During backpropagation, a denoising step $t'$ is randomly sampled from the $T$ reduced sampling steps, and gradients are retained \emph{only} for the $t'$-th denoising step of the $k$-th chunk, while all other steps are detached from the computational graph. This stochastic truncation yields a memory-efficient yet unbiased approximation of the full training objective, which can be expressed as:
\begin{equation}
\label{eq:stochastic_truncation}
\nabla_\theta \mathcal{L}
=
\mathbb{E}_{k \sim \mathcal{U}(1,K),\, t' \sim \mathcal{U}(1,T)}
\left[
\nabla_\theta \,
\mathcal{L}_{\text{DMD}}
\!\left(G^{(k,t')}_\theta(\mathbf{z})\right)
\right],
\end{equation}

where $G^{(k,t')}_\theta(\mathbf{z})$ denotes the model output at the $t'$-th denoising step of the $k$-th chunk, and all preceding chunks and denoising steps are detached from the computational graph during backpropagation.

Following this two-stage training strategy, SoulX-FlashTalk achieves state-of-the-art performance in both inference speed and generation quality compared to existing audio-driven video generation models.

\subsection{Real-time Inference Acceleration}

Purely optimizing training and inference in isolation is insufficient to fully satisfy stringent low-latency requirements. To achieve sub-second latency with the 14B-parameter model, we implement a full-stack acceleration suite specifically designed for 8-H800 node.

The core computational bottleneck lies within the DiT's massive attention operations. To dismantle this barrier, we deploy Hybrid Sequence Parallelism powered by xDiT. By synergizing Ulysses and Ring Attention mechanisms, we effectively distribute the attention workload, yielding a substantial 5$\times$ speedup in single-step inference compared to standard implementations. 
Furthermore, we optimize the DiT at the kernel level by adopting FlashAttention3, which is specifically designed to leverage the NVIDIA Hopper architecture, including its asynchronous execution pipeline. This enables improved overlap between data movement and computation, resulting in an additional 20\% reduction in attention latency compared to FlashAttention2.

As DiT inference becomes sufficiently accelerated, the computational overhead of the high-resolution VAE decoder emerges as the dominant latency factor. To address this paradigm shift, we introduce 3D VAE Parallelism to mitigate the decoding burden. By employing slicing-based strategies to distribute spatial decoding workloads across GPUs, we achieve an approximate 5$\times$ acceleration in VAE processing, ensuring that it does not become a pipeline bottleneck.

Finally, to eliminate overhead from the Python runtime and fragmented kernel execution, the entire inference pipeline is unified and optimized via \texttt{torch.compile}. This enables aggressive graph-level fusion and memory optimization, maximizing the hardware utilization limits of the H800 node.

\subsection{Architectural Analysis: Why Bidirectional?}
Although autoregressive models dominate streaming video generation, their inherent unidirectional dependency fundamentally constrains the modeling of global temporal structure. Under this paradigm, models primarily condition on historical frames and typically avoid strict frame-by-frame synthesis. 
Instead, generation is performed in minimal chunks to improve local consistency, where bidirectional attention is applied within each chunk, while unidirectional dependencies are enforced across chunks. 
However, this compromise remains insufficient to prevent temporal inconsistency, error accumulation, and identity drift, particularly in long-horizon generation.

We argue that, for the target task, incorporating long histories is not the primary bottleneck. Rather, effectively suppressing temporal drift and accumulated errors is of greater importance. 
Motivated by this observation, we fully preserve the bidirectional attention mechanism of the original model, consistently allowing all-to-all information exchange among frames. This design enables the model to jointly leverage past and implicit future context, resulting in more accurate and coherent generation at each step, while remaining fully aligned with the teacher architecture, thereby significantly accelerating model training.

Such bidirectional modeling not only substantially improves spatio-temporal coherence within individual chunks, but also provides a more robust and high-quality fundamental unit for streaming generation, thereby effectively mitigating drift and collapse issues in long-sequence video generation as a whole.

%% file: content_arxiv/3.Experiments.tex
\section{Experiment}
\label{sec:exp}
\textbf{Implementation.} We build our model upon the InfiniteTalk architecture, optimizing it to satisfy real-time constraints. The training pipeline begins with a lightweight Supervised Fine-Tuning (SFT) phase of $1,000$ steps to adapt the model to the reduced spatial resolutions and frame counts required for streaming. The subsequent distillation phase converges within $200$ steps.
We adhere to the Self-Forcing training paradigm, setting learning rates to $2 \times 10^{-6}$ for the Generator and $4 \times 10^{-7}$ for the Fake Score Network with a 1:5 update ratio. To simulate error accumulation in long-horizon generation, the Generator synthesizes up to $K=5$ consecutive chunks during distillation. To accommodate variable aspect ratios in real-world data, we employ a bucketing strategy across both SFT and distillation stages. All experiments utilize a cluster of 32 NVIDIA H20 GPUs with a per-GPU batch size of 1. We enable efficient training of the 14B-parameter model under these hardware constraints via Fully Sharded Data Parallel (FSDP)~\citep{zhao2023pytorch}, gradient checkpointing, and mixed-precision training.

\textbf{Datasets.} We source training and evaluation data from the public SpeakerVid-5M~\citep{zhang2025speakervid} and TalkVid~\citep{chen2025talkvidlargescalediversifieddataset} datasets, ensuring no overlap between training and test splits. To evaluate model performance, we construct a dedicated benchmark named TalkBench. This benchmark comprises two subsets: \textit{TalkBench-Short}, which contains 100 samples with durations under 10 seconds, and \textit{TalkBench-Long}, which consists of 20 samples exceeding 5 minutes.

\textbf{Evaluation Metrics.} We use the Q-Align visual-language model~\citep{wu2023q} for Image Quality Assessment (IQA) and Aesthetics Score Evaluation (ASE). Lip-audio synchronization is measured via Sync-C and Sync-D metrics~\citep{chung2016out}. Furthermore, we adopt VBench~\citep{huang2023vbench} to evaluate temporal quality, covering Subject Consistency (Subject-C), Background Consistency (BG-C), Motion Smoothness (Motion-S), and Temporal Flickering (Temporal-F).

\subsection{Performance of SoulX-FlashTalk}
SoulX-FlashTalk against state-of-the-art audio-driven generation models, including Ditto~\citep{li2025ditto}, EchoMimic-V3~\citep{meng2025echomimicv3}, StableAvatar~\citep{tu2025stableavatar}, OmniAvatar~\citep{gan2025omniavatar}, InfiniteTalk~\citep{yang2025infinitetalk}, and LiveAvatar~\citep{huang2025live}.

\subsubsection{Quantitative Analysis}
\input{table/table_main_results}

Table.~\ref{table:main_results} compares SoulX-FlashTalk with state-of-the-art methods on the TalkBench-Short and TalkBench-Long datasets.

On the short-video benchmark, SoulX-FlashTalk achieves the highest scores in visual quality and synchronization. Specifically, it records an ASE of $3.51$ and an IQA of $4.79$, surpassing the previous best performer, Echomimic-V3, which scored $3.45$ and $4.70$ respectively. In terms of lip-sync precision, our model attains a Sync-C score of $1.47$, outperforming the $1.32$ score of OmniAvatar. For inference speed, the system achieves a throughput of $32$ FPS on a 14B-parameter model. This exceeds the real-time requirement of $25$ FPS and significantly outperforms the $20.88$ FPS recorded by LiveAvatar.

Regarding temporal consistency metrics such as Subject-C and BG-C, Ditto records the highest values across both datasets, including a Subject-C of $99.80$. This performance is attributed to Ditto's generation paradigm which inpaints only the facial region while keeping the background and torso pixel-wise static. While this approach maximizes stability scores, it precludes the generation of full-body dynamics. SoulX-FlashTalk is designed to synthesize audio-driven full-body motion which naturally introduces greater pixel variance. Despite this increased complexity, it maintains a Subject-C score of $99.22$, demonstrating a balance between motion expressiveness and temporal stability.

For long-form generation, we assess robustness through synchronization retention. SoulX-FlashTalk achieves a Sync-C of $1.61$ and a Sync-D of $12.25$. These scores are better than InfiniteTalk and LiveAvatar. Additionally, the model sustains a throughput of $32$ FPS in long-duration tasks. These results confirm that the bidirectional distillation strategy effectively reduces the desynchronization and drift often found in unidirectional streaming models.

\subsubsection{Qualitative Analysis}

This section presents a qualitative assessment of SoulX-FlashTalk, focusing on generation fidelity, long-term stability, and lip-sync precision.
\input{figs/figure_visual_quality}

\textbf{Visual Fidelity and Detail Preservation.} Figure.~\ref{fig:visual_quality} compares the visual quality of 5-second video generations. Baseline models demonstrate significant struggles with plausible dynamics during large limb movements. Ditto fails to synthesize meaningful hand motion, with poses remaining static throughout the sequence, as highlighted in \textcolor{orange}{orange}. Echomimic-v3 and StableAvatar exhibit structural distortions and artifacts in hand regions, marked in \textcolor{blue}{blue}. InfiniteTalk suffers from hand over-exposure and excessive motion blur during rapid gestures. In contrast, SoulX-FlashTalk leverages its 14B DiT architecture and bidirectional attention mechanism to eliminate these artifacts. It synthesizes clear, structurally sound hand movements with sharp textures, avoiding the issues observed in baselines. Furthermore, our method surpasses LiveAvatar in background consistency and identity fidelity.

\input{figs/figure_long_term_stability}
\textbf{Stability in Infinite Generation.}
Figure.~\ref{fig:long_term_stability} evaluates generation stability over continuous sequences extending to 1000 seconds. Baseline models, including LiveAvatar, StableAvatar, and InfiniteTalk, suffer from significant error accumulation over time. As indicated by the \textcolor{blue}{blue} boxes, these methods exhibit severe texture blurring and loss of detail in background regions. SoulX-FlashTalk mitigates error propagation through bidirectional streaming distillation and its self-correction mechanism. As shown in the \textcolor{orange}{orange} boxes, our model preserves consistent facial geometry and sharp background details at the 1000-second mark, verifying its robustness for infinite streaming.

\begin{CJK*}{UTF8}{gbsn}
\textbf{Fine-grained Lip-sync Precision.} Figure.~\ref{fig:lip_sync} assesses lip-sync fidelity during specific Chinese phonetic articulations. Baseline methods struggle with complex phonemes, exhibiting structural misalignment. As highlighted by the \textcolor{yellow}{yellow} dashed boxes, during the articulation of characters such as “上” (shàng) and “突” (tū), competitors fail to match the mouth aperture and shape of the Ground Truth (GT), resulting in visible distortions. Conversely, SoulX-FlashTalk captures these fine-grained phonemic dynamics, yielding lip geometries that are rigorously aligned with the GT. This precision minimizes lip-sync drift and stiffness, ensuring visual authenticity across different languages.
\input{figs/figure_lip_sync}
\end{CJK*}

\input{table/table_ablation_chunk}
\subsection{Distillation Ablations} 
\subsubsection{Impact of Multi-step Retrospective Self-Correction}
We analyze how the number of generated chunks $K$ and scheduling strategies affect long-term stability. We compare fixed chunk strategies where $K$ equals $1$, $3$, or $5$ against a Random Strategy that samples $K$ from $1$ to $5$ during training.

Table.~\ref{tab:ablation_chunks} indicates that training with a single chunk $K=1$ yields the lowest training cost of $2.33$ hours but fails to maintain long-term stability. This is evidenced by a low Sync-C score of $1.12$ on long videos, confirming the issue of error accumulation. Increasing $K$ to $3$ improves stability significantly. However, further increasing $K$ to 5 raises the training cost to $6.40$ hours without delivering proportional gains in synchronization performance. The Random strategy achieves the best overall balance. It attains the highest Long Sync-C score of $1.61$ and optimal visual quality metrics while maintaining a moderate training cost of $4.40$ hours. This demonstrates that exposing the model to varying autoregressive lengths during distillation effectively improves robustness against accumulated errors.

\input{table/table_ablation_motion}

\subsubsection{Impact of Motion Latent Conditioning in DMD}
We examine the conditioning of the Real Score network across three dimensions: the source of motion latents, noise injection, and loss computation. Table.~\ref{tab:ablation_motion} shows that using student-predicted motion latents yields better visual quality than using Ground Truth (GT) latents. Specifically, the Predicted strategy with noise achieves an ASE of $3.51$ and an IQA of $4.79$, surpassing the GT configuration which scores $3.48$ and $4.77$. This indicates that using predicted latents helps reduce the disparity between training and inference.

Regarding noise and loss, injecting noise into predicted latents improves performance, raising ASE from $3.46$ to $3.51$. Conversely, including motion latents in the loss computation lowers the ASE to $3.48$. This suggests that requiring the model to reconstruct conditioning frames diverts focus from the primary denoising task. Thus, the configuration of Predicted Latents with Noise Injection and No Loss delivers the optimal results.

\subsection{Inference Latency Analysis}

\input{table/table_inference_speed}
In this section, we analyze component-wise latency on a single-node system with varying numbers of NVIDIA H800 GPUs. The experimental setup targets high-fidelity streaming at a resolution of $720 \times 416$ with 4-step denoising. Each clip comprises 33 frames, including 28 generated frames and 5 motion frames. Under this configuration, the pipeline achieves a throughput of up to 32 FPS.
\input{figs/figure_8gpus_latency}

The latency of the VAE and DiT is first examined to highlight the necessity of multi-GPU parallelism, as summarized in Table~\ref{table:diff_num_gpus}. On a single GPU, DiT inference alone incurs a latency of $1070$ ms per step, while VAE inference requires $97$ ms for motion frames encoding and $988$ ms for generated frames decoding. 

When scaling to 8 GPUs, DiT and VAE are parallelized using xDiT’s hybrid sequence parallelism and LightX2V’s slicing-based parallel strategy, respectively.
Due to inter-GPU communication overhead, the acceleration is slightly sub-linear, resulting in an overall speedup of nearly $5\times$.
Concretely, DiT latency is reduced from $1070$ ms to $193$ ms, VAE encoding from $97$ ms to $21$ ms, and decoding from $988$ ms to $192$ ms. Additional latency reductions are achieved by enabling \texttt{torch.compile}.

Building upon the core component optimizations, we report the end-to-end pipeline latency on an 8 $\times$ H800 GPU cluster in Figure.~\ref{fig:8gpus_latency}.
During the steady-state generation loop, the total latency per cycle is $876$ ms, of which audio processing takes $33$ ms, the core 4-step DiT denoising accounts for $616$ ms, frame decoding consumes $187$ ms, and motion-frame encoding requires $14$ ms. The remaining latency is attributed to miscellaneous overheads.
By achieving sub-second end-to-end latency, the proposed pipeline meets the stringent throughput requirements of real-time streaming.

%% file: table/table_main_results.tex
\begin{table*}[t]
\centering
\caption{Quantitative comparison of SoulX-FlashTalk and state-of-the-art methods on TalkBench-Short ($10$ s) and TalkBench-Long ($>5$ min). Models marked with $^*$ support real-time inference, and $^\triangle$ denotes the LoRA version.}
\label{table:main_results}
\resizebox{\textwidth}{!}{%
\begin{tabular}{lcccccccccccc}
\toprule
\multirow{2}{*}{\textbf{Dataset}} & \multirow{2}{*}{\textbf{Model}} & \multicolumn{9}{c}{\textbf{Metrics}} \\
\cmidrule(lr){3-11}
& & ASE $\uparrow$ & IQA $\uparrow$ & Sync-C$\uparrow$ & Sync-D$\downarrow$ & Subject-C $\uparrow$ & BG-C$\uparrow$ & Motion-S$\uparrow$ & Temporal-F$\uparrow$ & FPS $\uparrow$\\
\midrule
\multirow{6}{*}{Short} &
Ditto$^*$~\citep{li2025ditto} & 3.10 & 4.37 & 1.04 & 12.58 & \textbf{99.80} & \textbf{99.23} & \textbf{99.75} & \textbf{99.86} & \underline{21.80}\\
& Echomimic-V3~\citep{meng2025echomimicv3} & \underline{3.45} & \underline{4.70} & 0.89 & 12.81 & 98.75 & 95.98 & 99.54 & 99.44 & 0.53\\
& StableAvatar~\citep{tu2025stableavatar} & 3.05 & 3.01 & 0.78 & 12.12 & 98.34 & 96.46 & 99.44 & 99.06 & 0.64\\
& OminiAvatar~\citep{gan2025omniavatar} & 3.06 & 3.01 & \underline{1.32} & \underline{11.85} & 98.64 & 96.95 & 99.60 & \underline{99.66} & 0.16\\
& LiveAvatar$^*$~\citep{huang2025live} & 3.10 & 3.25 & 1.01 & 12.10 & 98.27 & 97.48 & 99.25 & 98.86 &  20.88\\
& Infinitetalk$^\triangle$~\citep{yang2025infinitetalk} & 3.09 & 3.04 & 1.25 & 11.89 & 98.71 & 97.04 & 99.54 & 99.34 & 5.10 \\
& SoulX-FlashTalk$^*$ (Ours) & \textbf{3.51} & \textbf{4.79} & \textbf{1.47} & \textbf{11.56} & \underline{99.22} & \underline{98.10} & \underline{99.61} & 99.52 & \textbf{32.00}\\
\midrule
\multirow{6}{*}{Long} &
Ditto$^*$~\citep{li2025ditto} & 3.11 & 3.03 & 0.83 & 13.45 & \textbf{99.80} & \textbf{99.23} & \textbf{99.76} & \textbf{99.89} & 0.53 \\
& StableAvatar~\citep{tu2025stableavatar} & 3.11 & 3.02 & 0.70 & 12.73 & 98.38 & 96.42 & 99.42 & 98.99 & 0.64 \\
& OmniAvatar~\citep{gan2025omniavatar} &2.92  &  2.82& 0.99 & 12.80 & 85.42 & 86.59 &  99.61& 99.57 & 0.16 \\
& LiveAvatar$^*$~\citep{huang2025live} & \textbf{3.15} & \textbf{3.06} & 0.96 & 12.73 & 98.19 & 97.39 & 99.26 & 98.87 & \underline{20.88}\\
& InfiniteTalk$^\triangle$~\citep{yang2025infinitetalk} & 3.12 & 3.03 & \underline{1.51} & \underline{12.28} & 98.97 & 97.32 & 99.59 & 99.50 & 5.10 \\
& SoulX-FlashTalk$^*$ (Ours) & \underline{3.12} & \underline{3.04} & \textbf{1.61} &\textbf{12.25 } & \underline{99.29} & \underline{98.36} & \underline{99.63} & \underline{99.58} & \textbf{32.00} \\
\bottomrule
\end{tabular}
}
\end{table*}

%% file: figs/figure_visual_quality.tex
\begin{figure*}[t]
    \centering
    \includegraphics[width=1.0\linewidth]{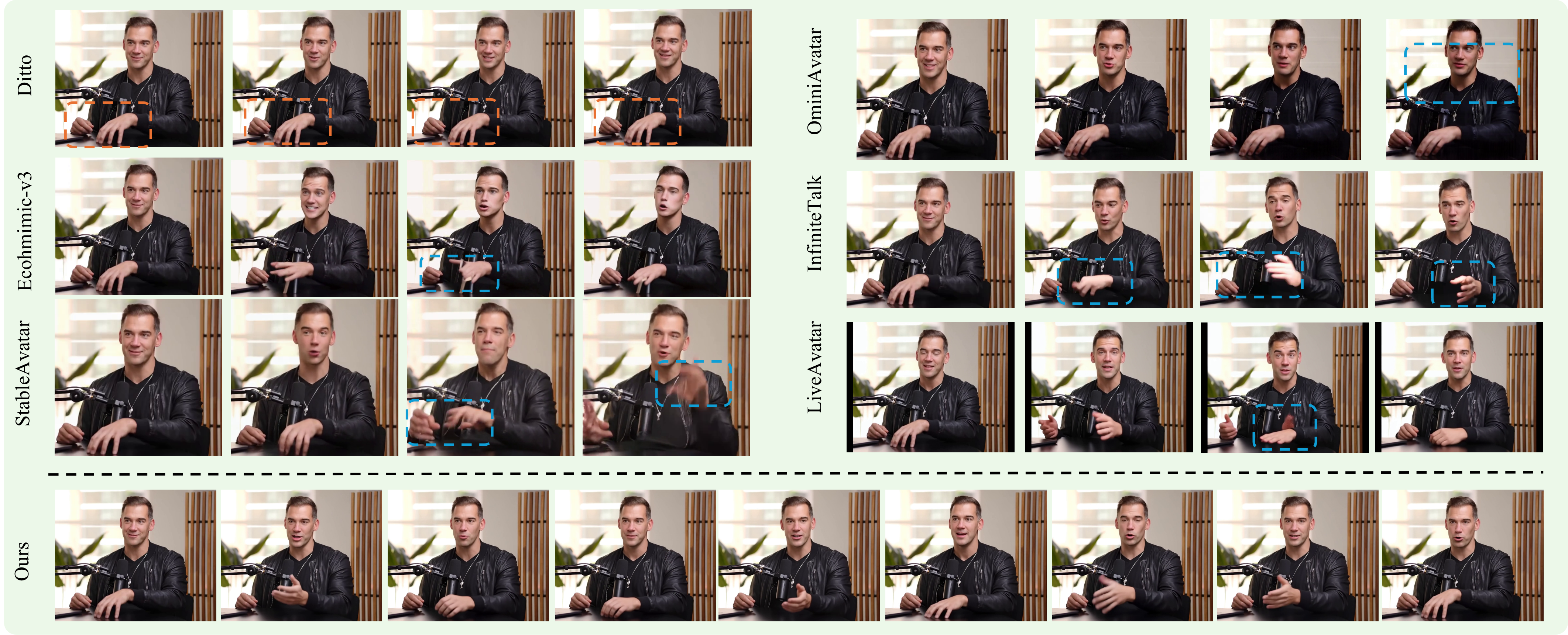}
    \caption{\textbf{Visual quality comparison on 5-second video generation.} \textcolor{orange}{Orange} boxes highlight static hand poses in Ditto, while \textcolor{blue}{blue} boxes highlight significant artifacts (e.g., hand distortion, over-exposure) in baselines. In contrast, SoulX-FlashTalk eliminates these artifacts, demonstrating superior structural integrity and detail fidelity.}
    \label{fig:visual_quality}
\end{figure*}

%% file: figs/figure_long_term_stability.tex
\begin{figure*}[t]
    \centering
    \includegraphics[width=0.8\linewidth]{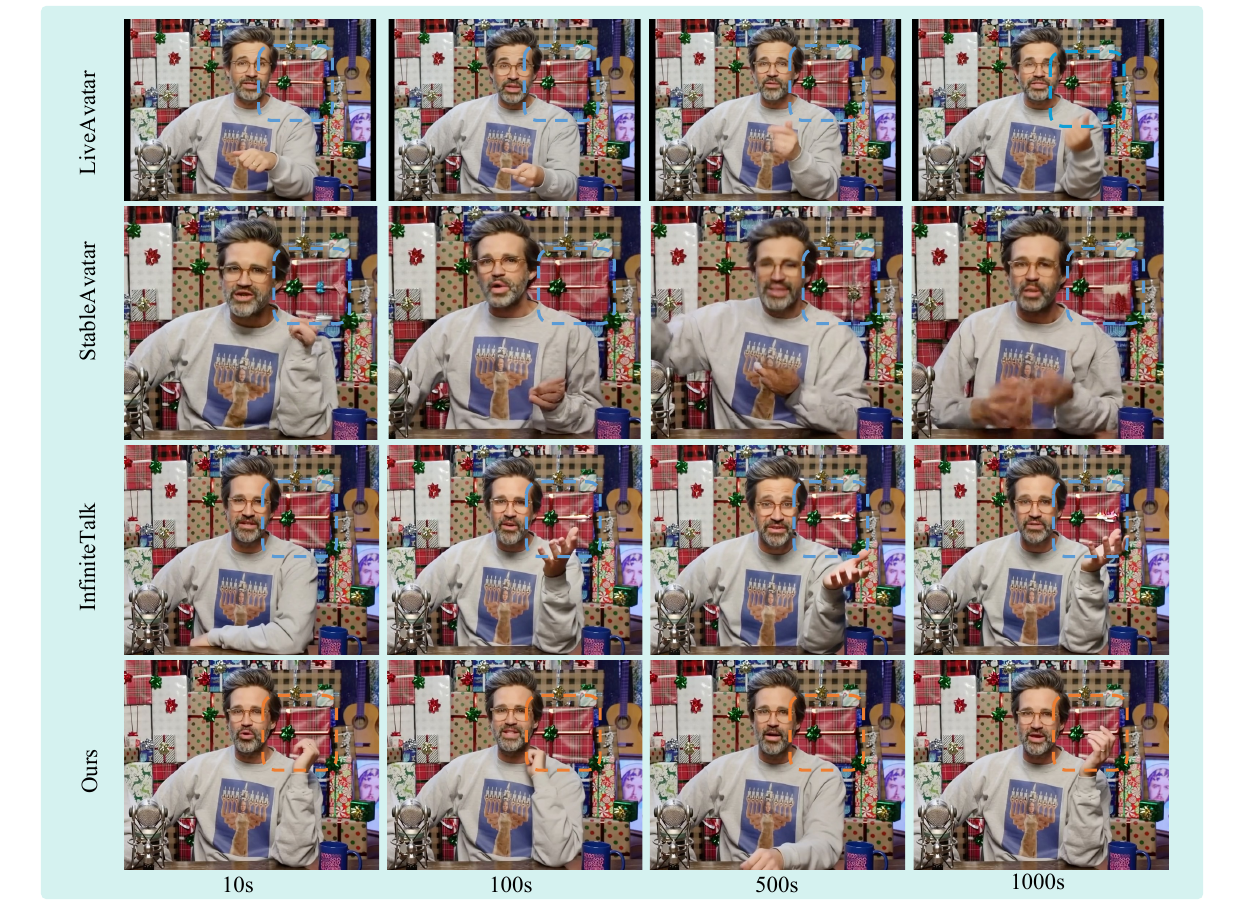}
    \caption{
    \textbf{Qualitative evaluation of long-term stability.} Comparison across $10$s to $1000$s reveals structural collapse in baselines (\textcolor{blue}{blue} boxes) versus the sustained robustness of SoulX-FlashTalk (\textcolor{orange}{orange} boxes), which preserves sharp details even after 1000 seconds of continuous generation.}
    \label{fig:long_term_stability}
\end{figure*}

%% file: figs/figure_lip_sync.tex
\begin{figure*}[h]
    \centering
    \includegraphics[width=1.0\linewidth]{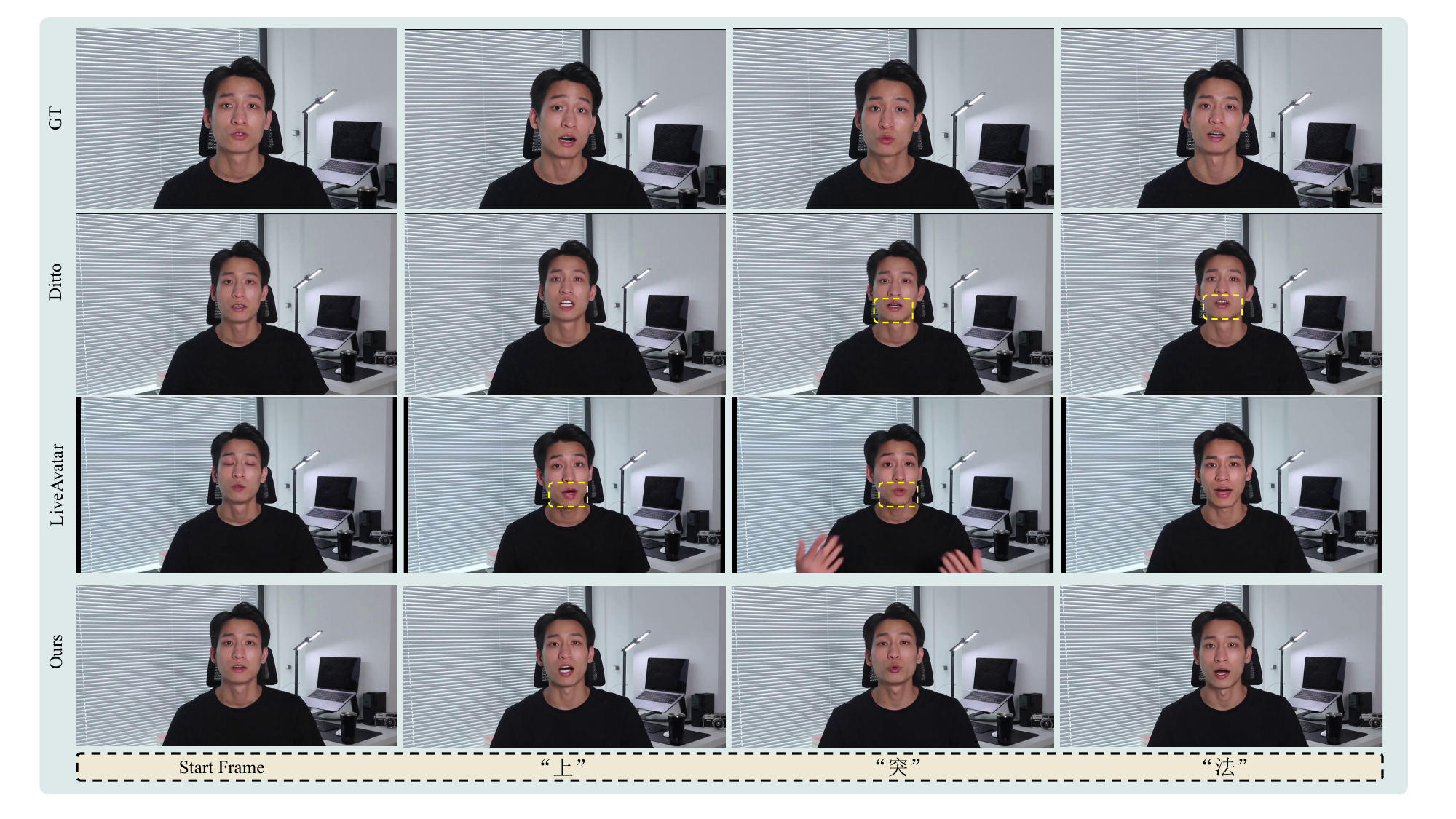}
    \caption{
    \textbf{Qualitative comparison of lip-sync precision on Chinese pronunciations.} \textcolor{yellow}{Yellow} dashed boxes indicate lip shape distortions in baseline methods for characters “上”, “突” and, “法” , whereas our method achieves high alignment with the Ground Truth (GT).}
    \label{fig:lip_sync}
\end{figure*}

%% file: table/table_ablation_chunk.tex
\begin{table}[h]
\centering
\caption{Ablation study on chunk generation strategies during distillation.}
\begin{tabular}{cc|cc|cc|cc|cc|c}
\toprule
\multicolumn{2}{c|}{Strategy} & \multicolumn{2}{c|}{ASE $\uparrow$} & \multicolumn{2}{c|}{IQA $\uparrow$} & \multicolumn{2}{c|}{Syn-C $\uparrow$} & \multicolumn{2}{c|}{Sync-D $\downarrow$} & \multirow{2}{*}{\makecell{Training\\Cost (h) $\downarrow$}} \\
& & short & long & short & long & short & long & short & long &\\
\midrule
\multirow{3}{*}{Fixed} & $K=1$  & 3.42 & 3.08 & 4.69& 2.99 & 1.39 & 1.12 & 11.79 & 12.47 & \textbf{2.33} \\
&$K=3$  & 3.47 & 3.09 & 4.75 & 3.00 & 1.39 & 1.59 & 11.69 & \textbf{12.02} & 4.40 \\
&$K=5$  & 3.50 & 3.10 & 4.79 & 3.01 & 1.35 & 1.44 & 11.87 & 12.33 & 6.40 \\
\midrule
Random & [1, 5] & \textbf{3.51} & \textbf{3.12} & \textbf{4.79} & \textbf{3.04} & \textbf{1.47} & \textbf{1.61} & \textbf{11.56} & 12.25 & 4.40 \\
\bottomrule
\end{tabular}
\label{tab:ablation_chunks}
\end{table}

%% file: table/table_ablation_motion.tex
\begin{table}[t]
\centering
\caption{Ablation on the conditioning of motion latents in the Real Score Network.}
\begin{tabular}{l|cc|cccc}
\toprule
\multirow{2}{*}{Motion Source} & \multicolumn{2}{c|}{Noise Strategy} & \multicolumn{4}{c}{Metrics} \\
& Add Noise & Calc Loss & ASE $\uparrow$ &IQA $\uparrow$ & Sync-C $\uparrow$&Sync-D $\downarrow$\\
\midrule
Ground Truth Latents & \ding{55} & \ding{55} & 3.46 & 4.75 & 1.43 & 11.68 \\
Ground Truth Latents & \checkmark & \ding{55} & 3.48 & 4.77 & 1.34 & 11.82 \\
\midrule
Predicted Latents & \ding{55} & \ding{55} & 3.46 & 4.75 & 1.43 & 11.68 \\
Predicted Latents & \checkmark & \ding{55} & \textbf{3.51} & \textbf{4.79} & \textbf{1.47} & \textbf{11.56}\\
Predicted Latents & \checkmark & \checkmark & 3.48 & 4.77 & 1.35 & 11.94 \\
\bottomrule
\end{tabular}
\label{tab:ablation_motion}
\end{table}

%% file: table/table_inference_speed.tex
\begin{table}[htbp]
\centering
\caption{Inference latency breakdown across varying GPU counts (Unit: ms).}
\label{table:diff_num_gpus}
\begin{tabular}{l|c|c|cc}
\toprule
\multirow{2}{*}{\textbf{GPUs}} & \multicolumn{2}{c|}{\textbf{VAE} } & \multicolumn{2}{c}{\textbf{DiT} }\\
& \textbf{Encode}  & \textbf{Decode}  & w/o torch.compile & w/ torch.compile\\
\midrule
\textbf{1} & 97 & 988 & 1070 & 800 \\
\textbf{2} & 69 & 690 & 620 & 490 \\
\textbf{4} & 39 & 350 & 313 & 261\\
\midrule
\multirow{2}{*}{\textbf{8}} & 21 & 192 & \multirow{2}{*}{193} & \multirow{2}{*}{\textbf{154}} \\
\textit{\quad-w/ torch.compile} & \textbf{14} & \textbf{187} & & \\
\bottomrule
\end{tabular}
\end{table}


%% file: figs/figure_8gpus_latency.tex
\begin{figure*}[t]
    \centering
    \includegraphics[width=1.0\linewidth]{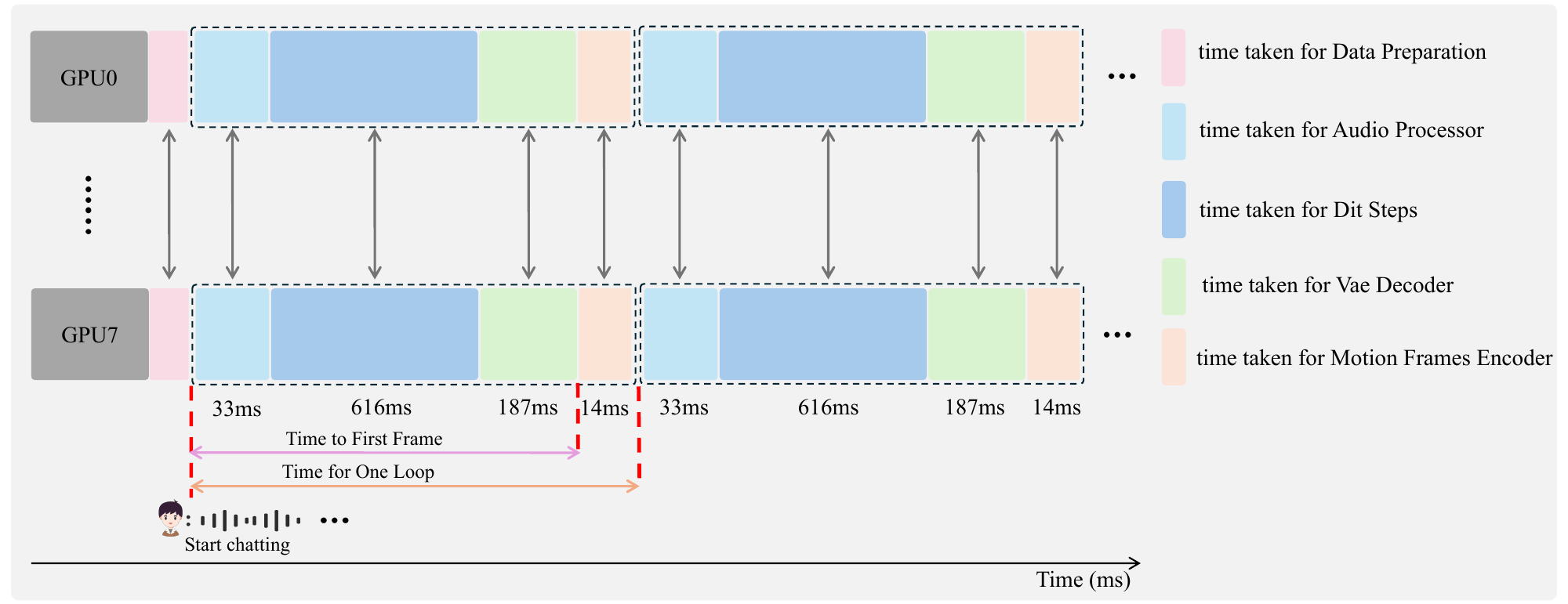}
    \caption{Detailed inference latency breakdown on 8 NVIDIA H800 GPUs.}
    \label{fig:8gpus_latency}
\end{figure*}

%% file: content_arxiv/4.Conclusion.tex
\section{Conclusion and Future Work}
We presented SoulX-FlashTalk, a framework designed to meet real-time requirements while maintaining high-quality video synthesis. Integrating Bidirectional Streaming Distillation with a Multi-step Self-Correction mechanism allows our 14B-parameter DiT model to sustain stable, infinite-length streaming on an 8$\times$H800 cluster. Our approach also simplifies training. We demonstrate that complex multi-stage pre-training is not required. A brief SFT phase followed by distribution matching distillation suffices to achieve state-of-the-art performance. We are open-sourcing this solution to serve as a practical baseline for the community.

Future work will prioritize model efficiency over system scaling. We intend to explore pruning, quantization, and optimized attention mechanisms. The objective is to deploy these models on consumer-grade hardware, eliminating the dependency on expensive computing clusters.

%% file: content_arxiv/6.Statement.tex
\section{Ethics Statement}
This research aims to advance digital human synthesis for beneficial applications. We confirm that all datasets utilized in this study are derived from publicly accessible academic repositories. The visual demonstrations presented in this report are fully synthetic and do not contain the Personally Identifiable Information (PII) of private individuals.

We acknowledge the dual-use nature of high-fidelity video generation technology and the potential risks associated with its misuse, such as the creation of deepfakes or the spread of misinformation. We firmly condemn any malicious application of this technology and advocate for the principles of Responsible AI. To mitigate these risks, we support the development of robust forgery detection algorithms and the implementation of invisible watermarking mechanisms to ensure content transparency and traceability. We remain committed to adhering to ethical guidelines and ensuring that our contributions promote the safe and positive evolution of the field.

%% file: content_arxiv/5.Contribution.tex
\section{Project Contributions}
All contributors are listed in no particular order.
\begin{itemize}
\item \textbf{Project Sponsor:} Ming Tao, Shunshun Yin
\item \textbf{Project Leader:} Siyuan Liu
\item \textbf{Algorithm:} Le Shen, Qian Qiao, Tan Yu
\item \textbf{Deployment \& Acceleration:} Ke Zhou, Tianhang Yu, Yu Zhan
\item \textbf{Data \& Evaluation:} Tan Yu, Tianhang Yu, Dingcheng Zhen
\end{itemize}